\documentclass[sigconf]{acmart}

\usepackage{bookmark}
\usepackage{amsfonts}
\usepackage{enumitem}
\setlist[itemize]{leftmargin=*}

\AtBeginDocument{%
  \providecommand\BibTeX{{%
    \normalfont B\kern-0.5em{\scshape i\kern-0.25em b}\kern-0.8em\TeX}}}

\setcopyright{acmcopyright}
\copyrightyear{2021}
\acmYear{2021}
\acmDOI{}

\acmConference[KDD '21]{KDD '21}{August 14--18, 2021}{Singapore}
\acmBooktitle{Workshop on Machine Learning in Finance (KDD ’21), August, 2021, Singapore}
\acmPrice{}
\acmISBN{}

\begin{document}

\settopmatter{printfolios=true}
\title{GuiltyWalker: Distance to illicit nodes in the Bitcoin network}

\author{Catarina Oliveira}
\email{catarina.a.oliveira@tecnico.ulisboa.pt}
\affiliation{%
  \institution{Instituto Superior Técnico}
  \country{Portugal}
}
\author{Jo\~ao Torres}
\email{joao.m.f.torres@tecnico.ulisboa.pt}
\affiliation{%
  \institution{Instituto Superior Técnico}
  \country{Portugal}
}
\author{Maria In\^es Silva}
\email{maria.silva@feedzai.com}
\affiliation{%
  \institution{Feedzai}
  \country{Portugal}
}
\author{David Apar\'icio}
\email{david.aparicio@feedzai.com}
\affiliation{%
  \institution{Feedzai}
  \country{Portugal}
}
\author{Jo\~ao Tiago Ascens\~ao}
\email{joao.ascensao@feedzai.com}
\affiliation{%
  \institution{Feedzai}
  \country{Portugal}
}
\author{Pedro Bizarro}
\email{pedro.bizarro@feedzai.com}
\affiliation{%
  \institution{Feedzai}
  \country{Portugal}
}


\begin{abstract}
    Money laundering is a global phenomenon with wide-reaching social and economic consequences. Cryptocurrencies are particularly susceptible due to the lack of control by authorities and their anonymity.  Thus, it is important to develop new techniques to detect and prevent illicit cryptocurrency transactions. In our work, we propose new features based on the structure of the graph and past labels to boost the performance of machine learning methods to detect money laundering. Our method, GuiltyWalker, performs random walks on the bitcoin transaction graph and computes features based on the distance to illicit transactions. We combine these new features with features proposed by \citet{weber2019antimoney} and observe an improvement of about 5pp regarding illicit classification. Namely, we observe that our proposed features are particularly helpful during a black market shutdown, where the algorithm by \citet{weber2019antimoney} was low performing. 
\end{abstract}




\maketitle

\section{Introduction}
\label{sec:introduction}

Money laundering is a serious financial crime that consists of the illegal process of obtaining money from criminal activities, such as drug or human trafficking, and making it appear legitimate. Cryptocurrencies, such as Bitcoin \cite{nakamoto2008peer}, are particularly susceptible to money laundering schemes due to their pseudo-anonymity and the relative lack of control by authorities. Preventing money laundering is an international effort and Anti-Money Laundering (AML) laws have been trying to cope with the new threats posed by criminals using cryptocurrencies \cite{union2018a, network2019a}.

In 2019, \citet{weber2019antimoney} released the Elliptic data set. It contains anonymized labeled Bitcoin transactions and enables researchers to study illicit behaviour in cryptocurrencies. The data set consists of a time-series graph with 200K labeled bitcoin transaction nodes and tabular data with 166 anonymized features describing each transaction. \citet{weber2019antimoney} assesses the performance of several supervised learning algorithms on the task of detecting nodes associated with illicit activities.

To improve existing supervised learning results found in the literature, we propose a new set of features that leverage the structure of the network and the existence of hubs or pockets of illicit transactions. We extract these new features with \emph{GuiltyWalker}. This random walker traverses a given network starting from a seed node and computes features related to the distance of the seed node to other nodes known to be illicit.

GuiltyWalker consists of two main components:

\begin{itemize}[leftmargin=0.5cm]
    \item \textbf{Random walker}: Given a transaction graph, a set of seed nodes, and the number of desired random walks for each of the seeds, GuiltyWalker samples random walks for each seed node. Due to the temporal nature of the graph, the walker only travels to past nodes (i.e., transactions) and stops at the first illicit node found or when there are no more valid nodes to visit.
    \item \textbf{Feature extractor}: Given a set of random walks for each of the seeds, GuiltyWalker computes aggregated features that summarize these walks, e.g., the average number of steps needed to reach one illicit node or the total number of different illicit nodes found.
\end{itemize}

In our experiments on the Elliptic data set, we observe that adding the features computed by GuiltyWalker improves the performance of machine learning methods; namely, we achieve a 5pp increased performance in F1-score, when compared against machine learning methods that use only the original anonymized features from \citet{weber2019antimoney}. Furthermore, the gains in performance are more pronounced during a black market shutdown, where the original performance by \citet{weber2019antimoney} dropped significantly.

This paper is organized as follows. Section 2 details the implementation of GuiltyWalker and how we generate new features from its output to enrich the data and convey additional information to supervised learning methods. Section 3 describes the experimental setup, and Section 4 the results consequently obtained. Section 5 presents the related work.  Finally, we set out the main conclusions in Section 6. 

\section{GuiltyWalker}
\label{sec:guiltywalker}

GuiltyWalker consists of a random walker that traverses a given transaction network from a seed node and extracts features based on the distance of that node to known illicit nodes. It includes two main components -- a random walker and a feature extractor, explained in the following subsections. 

\subsection{Random Walker}
\label{subsec:random_walker}

The random walker receives as input the original transaction graph $G$, a list of seed nodes, $S \subset V(G)$, and the number of \emph{successful} random walks desired, $k$. Successful random walks are explained later in this section.  GuiltyWalker's output for each seed node $s \in S$ is the list of sampled random walks $\mathcal{X}^s = \{X_{1}^s, X_{2}^s, ..., X_{k}^s\}$.

A random walk $X_i^s$ consists of a sequence of nodes
\begin{equation}
X_i^s = (x_1, x_2,...,x_n)\;,
\end{equation} such that $x_1 = s$. Due to the temporal nature of the transaction graph, the random walker can only walk backward in time. That is, it is only valid to go from node $x_n$ to node $x_m$ if $x_m$ represents a transaction older than $x_n$. This transaction network is represented as a directed graph connecting older nodes to newer nodes by an outgoing edge. Then, to address the former condition, during the random walk process, GuiltyWalker chooses a node uniformly at random from the incoming neighbors of the current node. When GuiltyWalker is in a given state of a random walk $X^s_i = (s, x_2,..., x_n)$, the process stops and returns $X^s_i$ as the final random walk if at least one of the following criteria is met:

\begin{itemize}[leftmargin=0.5cm]
    \item $x_n$ is a known illicit node/transaction.
    \item The set of eligible nodes to pick from is empty. This scenario happens when a given node has no incoming neighbors and, consequently, the random walker has no possible moves.
\end{itemize}

Otherwise, GuiltyWalker randomly picks the next node to add to $X^s_i$, and the process continues.

Note that since the edges only connect older transactions to newer transactions, there is always an end node in any random walk. In other words, the properties of our transaction graph guarantee that GuiltyWalker will not be stuck in an endless loop.

The number of \emph{successful} random walks, given by the user as input, intends to set the desired number of random walks ending in an illicit node from each seed node $s \in S$. However, as discussed before, the random walker may find a node with no incoming neighbors. In this case, a random walker finishes traversing the graph without reaching an illicit node. To ensure the number of desired successful random walks, GuiltyWalker performs as many random walks as needed, and only those are used for feature extraction. As we mention in Section~\ref{subsec:features_computation}, one of the features extracted from GuiltyWalker is the fraction of \emph{successful} random walks from the total number of random walks needed to perform to ensure the number of successful ones. This is a way of also considering the number of unsuccessful walks made from each node, which may be informative. 

It is important to note that some nodes in a transaction graph might have no paths to any illicit node. Thus, it is impossible to obtain successful random walks (as per our definition) for those nodes. To avoid this problem, GuiltyWalker first determines which nodes actually can reach an illicit node.
To do that, we first invert the direction of the graph. Then, we use the descendants algorithm for directed acyclic graphs from NetworkX \cite{SciPyProceedings_11} \cite{descendants}. It returns all nodes reachable from a source node $s$ in the graph $G$. Afterwards, we inspect if at least one of them is illicit. This procedure is made for all nodes in the transaction graph, and only those that can reach an illicit node are given as input for the random walker. 

 \subsection{Features Computation}
 \label{subsec:features_computation}

The second component of GuiltyWalker receives the list of random walks from each seed node and returns a data frame of features corresponding to each transaction, summarizing the random walks. In particular, GuiltyWalker obtains the following features:

\begin{itemize}[leftmargin=0.5cm]
    \item Minimum size of the random walks (min);
    \item Maximum size of the random walks (max);
    \item Mean size of the random walks (mean);
    \item Standard deviation of the random walks sizes (std);
    \item Median size of the random walks (median);
    \item First quartile of the random walks sizes (q25);
    \item Third quartile of the random walks sizes (q75);
    \item Fraction of \emph{successful} random walks from all the random walks performed by Random Walker (hit rate);
    \item Number of distinct illicit nodes in the random walks (illicit).
\end{itemize}

We also add information about the transaction nodes with no possible paths to fraudulent nodes to the data frame of features, with all features set accordingly (see Section~\ref{subsec:methodology}), due to the lack of information regarding the distance to an illicit node. 

\section{Experimental Setup}
\label{sec:experimental_setup}

\subsection{Elliptic Data Set}
\label{subsec:elliptic_data_set}

In this work, we use the Elliptic Data Set, a graph network of Bitcoin transactions\footnote[2]{\footnotesize Available at \url{https://www.kaggle.com/ellipticco/elliptic-data-set}}. Elliptic, a company focused on combating financial crime in cryptocurrencies, released this data set.

The data set includes 203,769 node transactions and 234,355 directed edges, representing the flow of Bitcoin currency (BTC) going from one transaction to the next. Each transaction can be categorized into three classes: "licit", "illicit" or "unknown", based on the category of the entity that created it. Licit categories include exchanges, wallet providers, miners, and financial service providers. Illicit categories include scams, malware, terrorist organizations, and Ponzi schemes. From the total number of transactions, 21\% (42,019) are labeled as licit, 2\% (4,545) as illicit, and the remaining 77\% (157,205) are unknown.

Besides the graph structure, the data set has 166 anonymized features associated with each transaction. The first 94 relate to information about the transaction itself, such as the time step, number of inputs/outputs, and transaction fee. The remaining features relate to aggregated information about the direct neighbors of the transaction, giving the maximum, minimum, standard deviation, and correlation coefficients of each transaction.

\begin{figure}[t!]
\center
\includegraphics[width=0.45\textwidth]{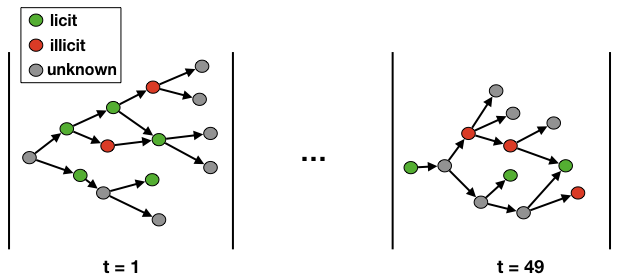}
\caption{Structure of the data set (taken from \cite{bellei_2019})}
\label{fig:structuredataset}
\end{figure}

Besides, a time step from 1 to 49 is associated with each node. It represents an estimate of when the Bitcoin network confirmed the transaction. The time steps are evenly spaced with an interval of about two weeks and each one contains a single connected component of transactions that appeared on the blockchain within less than three hours between each other. Therefore, it can be considered that this data set includes 49 directed acyclic graphs associated with different sequential moments in time. Figure \ref{fig:structuredataset} provides an idea of the structure of this data set. 

\subsection{Methodology}
\label{subsec:methodology}


This section gives an overview of the models used in our experiments and discusses our experimental setup. Following \citet{weber2019antimoney}, we perform a 70/30 temporal split of training and test data, respectively, for all experiments. Therefore, the train set includes all labeled samples up to the 34\textsuperscript{th} time step, and the test set includes all labeled samples from the last 15 time steps, up to the 49\textsuperscript{th}.

We use random forest for licit versus illicit prediction. First, we train the model on the train set using all 166 features and evaluate it on the entire test set. We use the scikit-learn \cite{pedregosa2011scikit} implementation of random forest, with 50 estimators, corresponding to the number of trees in the forest, and 50 max features, defined as the maximum number of features each tree can have. By doing so, we mimic the method in \citet{weber2019antimoney} enabling a fair comparison of the results. We also set the random state to 0 for the purpose of results reproducibility.

Then, we train a random forest model (using the same parameters as before) using (i) only the new set of features obtained by GuiltyWalker and (ii) both the features obtained by GuiltyWalker and the original 166 features. We extract the GuiltyWalker features after performing 100 \emph{successful}  random walks. Missing values for the transaction nodes that cannot reach an illicit node are filled with -1 values, except feature \textit{hit}, which is filled with 0 values, as it represents the fraction of random walks ending in a fraudulent node. We see that the utilization of some of these alternative sets of features improves performance in Section \ref{sec:results}.

To further improve the results, we filter the set of features obtained by GuiltyWalker to keep just the most important ones. This method characterizes every single feature's importance as the decrease in the performance score after randomly shuffling its position in the set, and is called Permutation feature importance \cite{featureimportance}. After applying this method and assessing every feature's importance, the new features kept for further classification purposes are \textit{hit}, \textit{std}, \textit{illicit}, \textit{max} and \textit{mean}. We also analyse the model's performance with these features together with the 166 original ones in Section \ref{sec:results}.

Similarly to \citet{weber2019antimoney}, we evaluate the random forest classifier's performance with each set of features using the F1-score for the illicit class, hereafter referred to as illicit F1-score. This score is the weighted average of precision and recall. Moreover, it is suitable for imbalanced tasks, which is the case of our dataset (91\% of licit nodes and 9\% of illicit ones). We also use the ROC curve (and AUC value) and precision and recall measures to evaluate the models' performance.

\section{Results}
\label{sec:results}

In this section, we present the results obtained by using the standard model, random forest, with the 166 baseline features (referred to as AF), as well as only the new features extracted from GuiltyWalker (referred to as GWF) and the former ones together with the latter (referred to as AF+GWF). Furthermore, we show the results obtained using the 166 features in conjunction with the new ones obtained after performing feature reduction 
(referred to as AF+GWF*). 

\begin{table}[ht]
\caption{Illicit classification results using random forest, for different features. {\normalfont \textit{AF}} refers to the original all features, {\normalfont \textit{GWF}} refers features extracted from the GuiltyWalker, and, {\normalfont \textit{GWF*}} refers to GuiltyWalker features after feature selection.}
\label{tab1:f1score}
\vspace{5mm}
\resizebox{\columnwidth}{!}{%
\renewcommand{\arraystretch}{1.5}
\begin{tabular}{lcccc}
\toprule
\multicolumn{1}{c}{}       &           & Illicit  & \multicolumn{1}{c}{}   & MicroAVG \\ \cline{2-4}
\multicolumn{1}{c}{Method} & Precision & Recall  & \multicolumn{1}{c}{F1} & F1       \\ \midrule
AF             & 0.91      & 0.72    & 0.80                    & 0.977    \\
GWF        & 0.93      & 0.11    & 0.20                    & 0.942    \\
AF + GWF        & 0.93      & 0.76    & 0.84                    & 0.981    \\
AF + GWF*     & 0.97      & 0.77    & 0.85                    & 0.983   \\
\bottomrule
\end{tabular}
}
\end{table}

Table \ref{tab1:f1score} shows the testing results in terms of precision, recall and F1-score concerning the illicit class. For the sake of completeness, we also show the micro-averaged F1 score.

An important thing to note from Table~\ref{tab1:f1score} is that the GuiltyWalker features alone are not informative enough. The F1-score value obtained using only these features is very low (0.20). We can also observe higher precision, recall, and F1-score when using GuiltyWalker's additional features, suggesting the importance of the graph structure. Using GuiltyWalker features, we improved precision, recall, and F1-score by 2 percentage points (pp), 4pp, and 4pp, respectively.

In order to understand the importance of each one of the features created, we performed feature importance, using the method described in the previous section. We kept only the most important features to train together with the original ones. Results show that by filtering GuiltyWalker features and keeping only the most important ones (\textit{hit}, \textit{std}, \textit{illicit}, \textit{max} and \textit{mean}), the performance of the model slightly improves (we improved F1-score by 1pp, comparing with the model AF+GWF).

To give additional insights about the performance of the new model AF + GWF* compared against the original model, we plot the ROC curve of both models. 

\begin{figure}[t!]
\center
\includegraphics[scale=0.17]{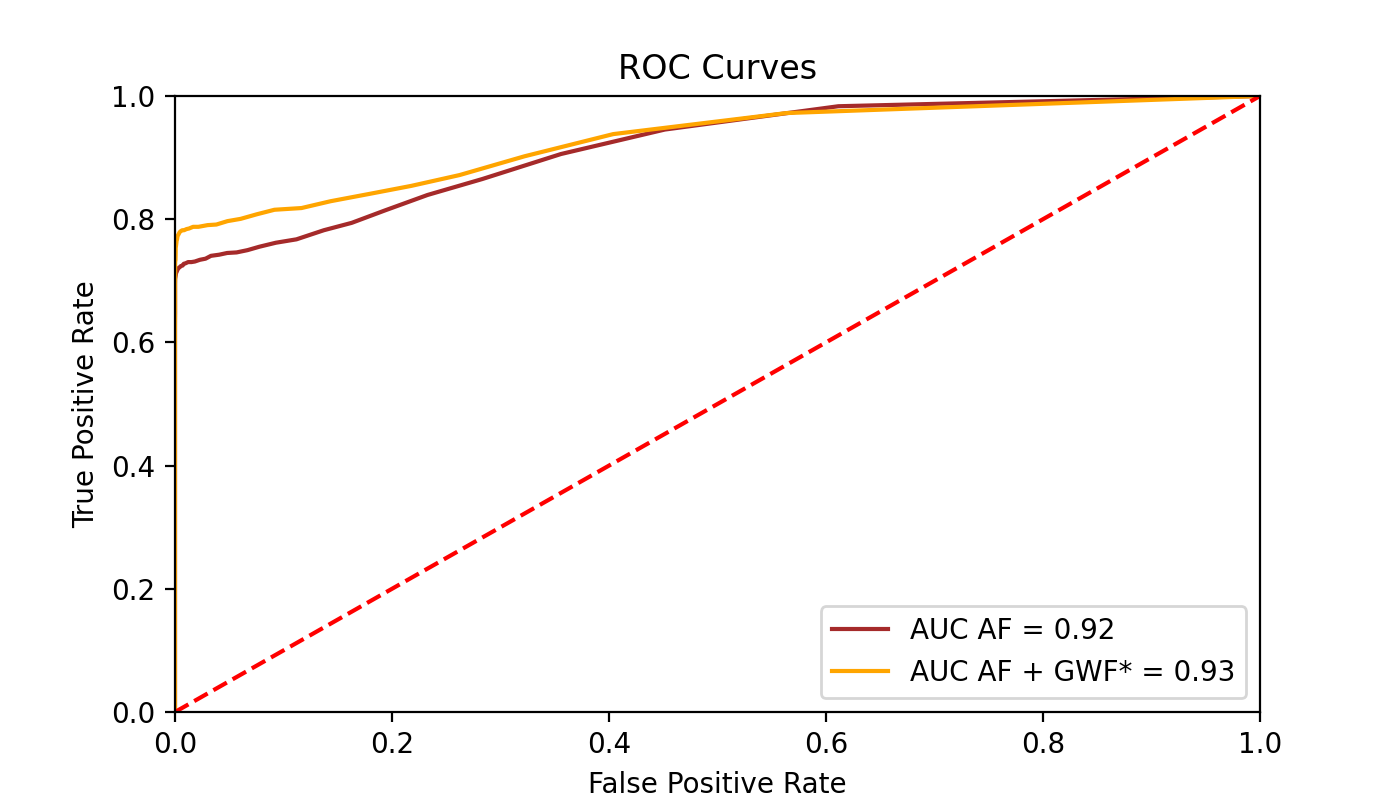}
\caption{ROC curve associated with AF + GWF* and AF models. Random baseline represented in red. }
\label{fig:roc}
\end{figure}

Note that the ROC curve shows the trade-off between sensitivity/ recall and specificity. Moreover, the area under the curve (AUC) can be seen as a measure of separability. In other words, it represents how much a model is capable of distinguishing between classes. Therefore, from the observation of Figure \ref{fig:roc}, we can infer that both models are quite good at predicting illicit nodes as illicit and licit ones as licit. However, AF + GWF* is slightly better (improves AUC value by 1pp). In particular, for very low false positive rates, our method seems to be significantly better. 

In a real scenario, we would be more interested in low false-positive regions of the ROC-curve since raising too many alerts is not practical. With this in mind, we compare the recall at specific low false positive rates, namely 1\%, 5\% and 10\%, and AF+GWF* shows considerable gains when compared against AF: recall@1\% increases from 73\% to 78\% (5pp), recall@5\% increases from 75\% to 80\% (5pp), and recall@10\% increases from 76\% to 82\% (6pp). Therefore, while the gain of using GuiltyWalker's features is only 1pp in the full region, in the region of interest the gain is considerably higher.

\begin{figure}[ht]
\center
\includegraphics[scale=0.17]{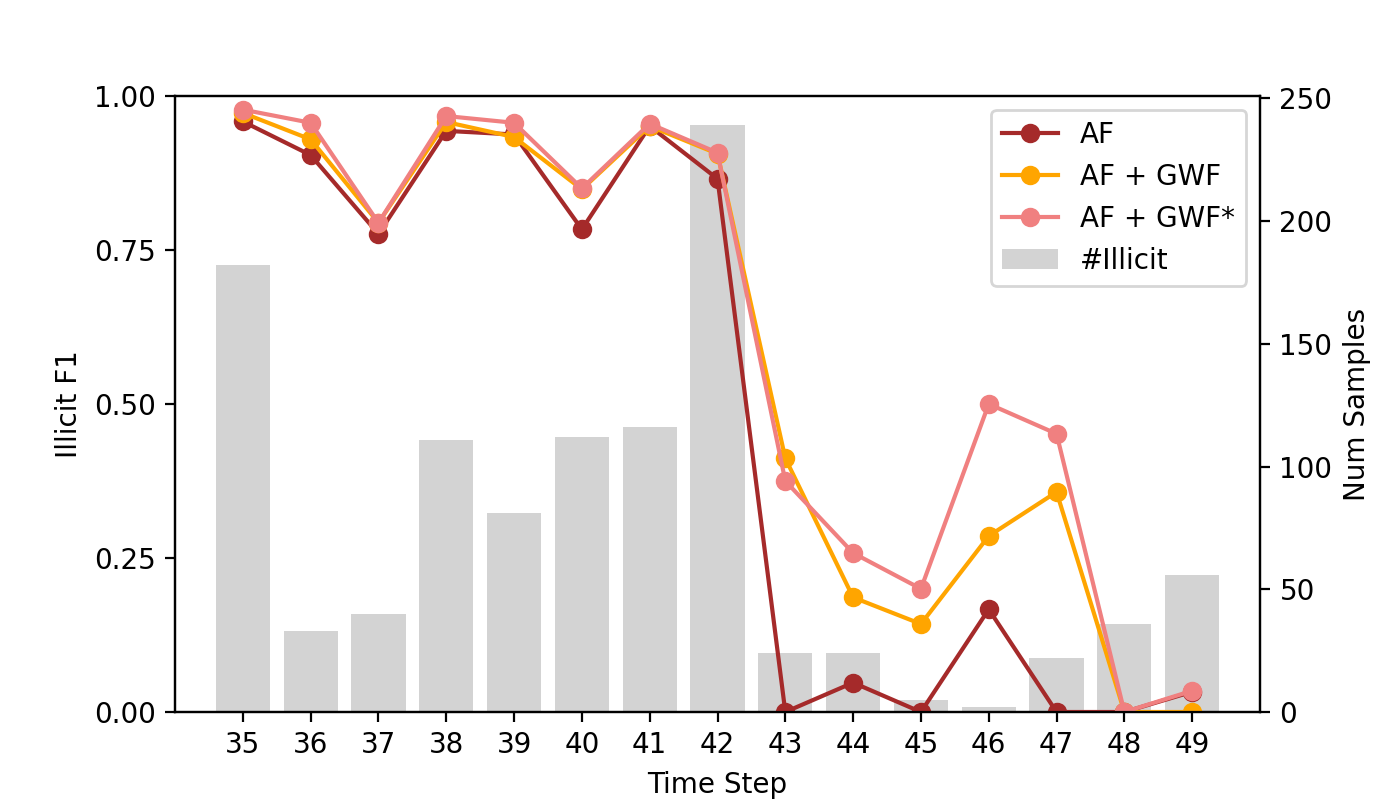}
\caption{Illicit F1-score obtained with Random Forest, for the standard 166 features and the new GuiltyWalker features before and after feature selection.}
\label{fig:fsgwf}
\end{figure}

As noted by \citet{weber2019antimoney}, a sudden dark market shutdown occurring at time step 43 extremely affects the model performance. In particular, the random forest model trained on the 166 features, from that time step forward, cannot achieve an Illicit F1-score value above 0.25. The introduction of the new set of features extracted from GuiltyWalker improves F1 results in the entire test set (i.e., time steps 35-49). However, this improvement is more notorious after this dark market shutdown (from time step 43 to 49). In fact, from time step 43 to time step 49, we observe, on average, a F1 score improvement of about 10pp and 16pp with AF + GWF and AF + GWF* models, respectively. Note that for time steps 48 and 49, both of these models still perform poorly.

As we can see in Figure \ref{fig:fsgwf}, both AF + GWF and AF + GWF* models are able to reliably capture new illicit transactions after the dark market shutdown, in comparison with the original model. To understand the additional information those models are capturing, we compute the confusion matrices of the models AF, AF+GWF and AF+GWF*. We obtain 784, 828 and 831 true positives (referred to as TP), respectively. We also determine the new TP found and the ones lost, using AF + GWF and AF + GWF* models, in comparison to the ones found training RF with the original set of features. By doing so, we can verify that with the new sets of features, AF+GWF and AF+GWF*, we found 48 and 50 new TP and lost 4 and 3 TP that the original model could find, respectively.

Concerning the AF + GWF* model, we observe that, for almost all new TP found, the features extracted from GuiltyWalker (\textit{max}, \textit{mean}, \textit{std}, \textit{illicit} and \textit{hit}) have positive values. Only 2 of the 50 elements have -1 values regarding \textit{max}, \textit{mean}, \textit{std} and \textit{illicit} and 0 with respect to \textit{hit}. 
Recall that these values indicate that the associated transaction nodes have no possible paths to known illicit nodes.

This interesting fact lets us infer that the new set of features adds some new information based on the graph's structure, which allows the model to make better predictions. However, we have to notice that the fact that a given node has a path to an illicit transaction does not necessarily imply that it is also illicit and vice-versa. This information alone is not enough to make good predictions concerning the labels of the transaction nodes, as we verified from the results obtained for the GWF model. Nonetheless, it provides extra information to complement the original features in a way that boosts performance of the overall model.

\section{Related Work}

Besides the work of \citet{weber2019antimoney}, which was the baseline for our study, more recently, \citet{lorenz2020machine} proposed active learning techniques to study the minimum number of labels necessary to achieve high detection of illicit activity in cryptocurrencies and tested them also on the Elliptic data set. Thus, even though using a different approach to the same problem, the authors did not aim to achieve better results than the baseline.  Moreover, \citet{alarab2020comparative} proposed an ensemble learning method, using a combination of the given supervised learning models, and applied it on the Elliptic data set, improving the baseline results. Although they improved upon existing results, our results, using the new set of features, are better. While \citet{alarab2020comparative} achieves higher precision than us (97.38\% versus 96.5\%), we achieve higher recall (76.7\% versus 72.2\%), higher F1-score (85.47\% versus 82.92\%), and higher accuracy (98.3\% versus 98.06\%). 

As far as we know, previous work on the application of graph-related features and, in particular, 
random walks, in a supervised learning setting are scarce. \citet{Hu2019CharacterizingAD} worked with Bitcoin transaction graphs and used various graph characteristics to differentiate money laundering transactions from regular transactions. They actually found that the main difference between them lies in their output values and neighbourhood
information. The authors also evaluated a set of classifiers based on different types of extracted features, namely immediate neighbours, curated features, deepwalk embeddings \citep{perozzi2014deepwalk}, and node2vec embeddings \citep{grover2016node2vec} to classify money laundering and regular transactions. This approach differs from ours as we are not trying to embed the graph or a particular node's neighbourhood but instead to describe distances to a specific target (i.e., malicious activity). Nonetheless, the descriptive power of random walks in networks is still recognized. 
\citet{socialnetwork} studied methods based on the iterative application of traditional classifiers using graph information as features, and methods that propagate the existing labels via random walks. Moreover, concerning the application of random walks in the context of classification problems, \citet{textclassification} proposed a new approach for estimating term weights in a document based on a random walk model. They showed that the new random walk based approach outperforms the traditional term frequency approach of feature weighting. Therefore, with this work, we extend the existing knowledge regarding random walks to improve classifiers' performance in graph datasets. 

\section{Conclusion}
\label{sec:conclusion}

In this study, we set out to improve the performance of supervised models in an anti-money laundering classification task. Given a transaction network, we propose a method called GuiltyWalker that extracts information from the structure of the network and the existence of past labels to create new features for a supervised model. It consists of a random walker that traverses the transaction network starting from a seed node and a feature extractor that computes features related to the distance of the seed node to other nodes known to be illicit.

We test our method on a public dataset of Bitcoin transactions published by \citet{weber2019antimoney}. Using a supervised setting similar to the original authors as our baseline, we showed that by training the same classifier considering the original 166 features and the new ones extracted from GuiltyWalker, we could obtain better results. In particular, by filtering the features extracted from GuiltyWalker and considering only the most important ones, the results were even better. The performance differences were more notorious for time steps associated with a black market shutdown, where the baseline model performed poorly. Moreover, we observed that the models that considered GuiltyWalker features could reliably capture new illicit transactions that were not captured by the model from \citet{weber2019antimoney}.


\bibliographystyle{ACM-Reference-Format}
\bibliography{references}

\end{document}